\documentclass[preprint,12pt]{elsarticle}

\usepackage{amssymb}
\usepackage{amsmath,amsfonts}
\usepackage[numbers]{natbib}
\usepackage[linesnumbered,ruled,vlined]{algorithm2e}
\usepackage{algorithmic}
\usepackage{array}
\usepackage{subfigure}
\usepackage{textcomp}
\usepackage{stfloats}
\usepackage{url}
\usepackage{verbatim}
\usepackage{graphicx}
\usepackage{hyperref}
\usepackage{multirow}
\usepackage{makecell}
\usepackage{lscape}

\journal{-}

\begin{document}

\begin{frontmatter}

\title{Neighborhood-Order Learning Graph Attention Network for Fake News Detection}

\author[aut]{Batool Lakzaei}
\ead{b\_lakzaei@aut.ac.ir}

\author[aut]{Mostafa Haghir Chehreghani\corref{cor1}}
\ead{mostafa.chehreghani@aut.ac.ir}

\author[aut]{Alireza Bagheri}
\ead{ar\_bagheri@aut.ac.ir}

\affiliation[aut]
{	organization={Department of Computer Engineering, Amirkabir University of Technology (Tehran Polytechnic)},            
	city={Tehran},
	postcode={1591634311},             
	country={Iran}}

\cortext[cor1]{Corresponding author}

\begin{abstract}
Fake news detection is a significant challenge in the digital age, which has become increasingly important with the proliferation of social media and online communication networks.
Graph Neural Networks (GNN)-based methods have shown high potential in analyzing graph-structured data for this problem.
However, a major limitation in conventional GNN architectures is their inability to effectively utilize information from neighbors beyond the network’s layer depth, which can reduce the model's accuracy and effectiveness.
In this paper, we propose a novel model called Neighborhood-Order Learning Graph Attention Network (NOL-GAT) for fake news detection. This model allows each node in each layer to independently learn its optimal neighborhood order.
By doing so, the model can purposefully and efficiently extract critical information from distant neighbors.
The NOL-GAT architecture consists of two main components: a Hop Network that determines the optimal neighborhood order and an Embedding Network that updates node embeddings using these optimal neighborhoods.
To evaluate the model's performance, experiments are conducted on various fake news datasets. Results demonstrate that NOL-GAT significantly outperforms baseline models in metrics such as accuracy and F1-score, particularly in scenarios with limited labeled data.
Features such as mitigating the over-squashing problem, improving information flow, and reducing computational complexity further highlight the advantages of the proposed model.
\end{abstract}


\end{frontmatter}

\section{Introduction}
\label{sec:intro}
In today's world, the Internet and social media platforms play a central role in communication and information sharing, with social networks becoming the primary channels for interaction \cite{paraschiv2022unified}, such that many individuals prefer online resources and social networks over traditional media for communication, entertainment, and news \cite{xie2023heterogeneous}.
The unreliability of information disseminated on these platforms has drawn attention to the spread of misinformation on social networks as a significant social issue.
A series of incidents in recent years have demonstrated that substantial harm, in the form of social disruption, can arise due to fake news \cite{choudhary2024gin}, as fake news contains false and misleading information presented in the format of a news story, thereby possessing the potential to cause widespread turmoil across the globe \cite{teja2023graph}.

In recent years, researchers have proposed various methods for detecting fake news using machine learning techniques, particularly deep learning approaches, which are primarily based on supervised learning methods. 
These approaches often face challenges such as the lack of large, labeled datasets, making them costly and time-consuming \cite{golo2023one}. 
Semi-supervised methods, which rely on a small amount of labeled data, can provide an effective solution to overcome these challenges.

The representation of data plays a critical role in the success of machine learning algorithms, particularly in scenarios with a scarcity of labeled data \cite{van2020survey}. 
Graph-based models efficiently analyze complex and non-Euclidean data \cite{breve2011particle}~and excel in semi-supervised learning by capturing structural dependencies~\cite{shi2020skeleton}. 
These models have demonstrated high effectiveness for semi-supervised learning by leveraging structural and contextual relationships between data points~\cite{rossi2016optimization}. 
However, traditional machine learning methods and neural networks are not well-suited for processing graph data due to the non-Euclidean nature of such data.
Graph Neural Networks (GNNs)~\cite{DBLP:conf/iclr/KipfW17,DBLP:conf/iclr/XuHLJ19,10.1145/3700790,DBLP:conf/icml/YouYL19,DBLP:journals/tjs/ZohrabiSC24,DBLP:conf/complexnetworks/LuanHLZCP23}, introduced in recent years, have addressed this challenge with their unique capability to process graph data. 
These models enable effective learning from non-Euclidean data by employing specialized architectures designed to capture the complex structures and nonlinear relationships inherent in graphs.
In Graph Neural Networks, the goal is to learn a function that maps each node of the input graph to a low-dimensional vector space. 
This mapping must possess the property of similarity preserving. 
In other words, if two nodes share similar features and structural roles within the graph, they should be mapped to nearby points in the vector space. 
The generated vector for each node is referred to as the embedding of that node. 
These embeddings can be used as inputs (serving as feature vectors) in various machine learning algorithms \cite{haghir2022half}.
GNNs not only facilitate better analysis of graph data but have also demonstrated exceptional performance across a wide range of applications, including semi-supervised learning \cite{de2022network}.
They have been successfully applied to tasks such as object detection \cite{shi2020point}, recommendation systems \cite{gao2022graph,DBLP:journals/corr/abs-2402-03365}, and sentiment analysis \cite{lu2022aspect}. 
In fake news detection, GNNs outperform other methods by integrating content and contextual features into a unified framework \cite{hu2019multi, monti2019fake, ren2020adversarial, benamira2019semi}. 
The use of GNNs for fake news detection provides a powerful solution to mitigate the adverse effects of misinformation and preserve the integrity of online information sharing \cite{hiremath2023analysis}.

Despite the remarkable success of graph neural networks in various fields, including fake news detection, there are still challenges related to message propagation within these networks. In the standard GNN architecture, in each layer, the embedding vector of each node is updated based on the vectors of its neighbors. In an $L$-layer network, each node can access at most its neighbors up to the $L$-hop order, meaning that valuable information from nodes further away will be inaccessible. This limitation can lead to decreased model performance in many practical applications.

In this paper, we propose a novel method called
Neighborhood-Order Learning Graph Attention Network (NOL-GAT), aimed at addressing this challenge. In the proposed method, each node independently and adaptively learns the optimal neighborhood order for updating its embedding vector in each layer. Unlike conventional methods that limit the node’s neighborhood to a maximum of $L$ layers, this capability allows the model to leverage more effective and flexible information.

With this approach, each node can learn the optimal neighborhood order tailored to the graph structure and local data features, without relying on pre-determined settings. This not only improves the model’s accuracy but also enables access to key information in nodes further away, while still preventing issues such as over-smoothing.

Our contributions can be summarized as follows:
\begin{itemize}
	\item \textbf{Introducing a novel adaptive architecture for graph neural networks:}
	
	We propose the Neighborhood Learning Graph Attention Network (NOL-GAT), a new architecture that enables each node to independently select its optimal k-hop neighbors in every layer.
	This feature enhances the model's flexibility in learning from both close and distant neighbors.
	We leverage the Gumbel-Softmax estimator to learn the categorical distribution for neighbor's order selection in a differentiable manner, which enhances optimization and performance.
	Our approach integrates two key components:  
	\begin{itemize}
		\item Hop Network ($\Phi$): Learns the optimal neighborhood order for each node.
		\item Embedding Network ($\Psi$): Updates node embeddings based on the optimal neighbors.		
	\end{itemize}
	This hybrid design improves adaptability and model efficacy.	
	
	\item \textbf{Addressing core challenges in graph neural networks:}
	\begin{itemize}
		\item Over-smoothing mitigation: By selectively choosing neighbors at each layer, NOL-GAT prevents the excessive homogenization of node features. 
		\item Reducing over-squashing: The model efficiently propagates important information from distant nodes without overwhelming the target node with compressed information. 
		\item Lower computational complexity: The adaptive neighbor selection reduces the need for deep layers or complex message-passing mechanisms.
	\end{itemize}

	\item \textbf{Independent decision-making across nodes and layers:}
	
	Each node independently determines its neighbors based on local requirements, and this decision-making process remains layer-wise independent. This capability allows the model to adapt to diverse graph structures effectively.
	
	\item \textbf{Achieving superior results on real-world datasets:}
	
	The NOL-GAT model demonstrates state-of-the-art performance in fake news detection across five datasets. It significantly outperforms baseline models in terms of accuracy and macro-F1 score, especially under low-label conditions (10\%-30\% labeled data).	
\end{itemize}

The remainder of this paper is organized as follows.
Section \ref{sec:rw} reviews the related works in the field, providing a foundation for understanding the context of our research. In \ref{sec:problem_def}, we define the problem that motivates our study.
Section \ref{sec:proposed_model} presents the proposed model, detailing its design and underlying principles.
The properties and theoretical underpinnings of the model are discussed in
Section \ref{sec:model_properties}.
In Section \ref{sec:exp}, we describe the experiments conducted to evaluate the model's performance, followed by an analysis of the results. Finally, \ref{sec:conclusion} concludes the paper.


\section{Related Works}
\label{sec:rw}
Researchers have explored various techniques like natural language processing, machine learning, and social media analysis for fake news detection. 
These methods are generally divided into three categories \cite{lakzaei2024disinformation}:
\begin{itemize}
	\item \textbf{Content-based methods:}
	which focus on analyzing the text or visuals to identify fake news~\cite{lakzaei2024loss}.
	\item \textbf{Context-based methods:}
	which use contextual data such as user profiles and propagation patterns~\cite{DBLP:conf/bigdataconf/AsghariCC22}.
	\item \textbf{Hybrid methods:}
	which combine both content and context to detect fake news~\cite{DAVOUDI2022116635}.
\end{itemize}
Our approach is content-based, relying solely on textual content, which is the key feature in all existing datasets. As a result, our method can be applied to any dataset.
We note that context information may not always be available.

A significant portion of existing methods for fake news detection relies on supervised approaches \cite{yadav2024emotion, qu2024qmfnd, fang2024nsep}.
The primary challenge and limitation of this approach is its dependency on large, labeled datasets, which makes it time-consuming and costly to generate. 
Additionally, obtaining sufficient labeled data can be challenging, especially when it comes to handling the vast amounts of news data available.  
Therefore, semi-supervised approaches, which can work with a small subset of labeled data and classify the larger portion of unlabeled data, provide an effective solution. 
In this paper, we propose a semi-supervised approach based on Graph Neural Networks (GNNs). 
As a result, our focus is primarily on recent semi-supervised methods and those utilizing GNNs.

DEFD-SSL \cite{al2023robust} is a semi-supervised approach for fake news detection that combines various deep learning models, data augmentations, and distribution-aware pseudo-labeling. It uses a hybrid loss function and incorporates ensemble learning along with distribution alignment to maintain balanced accuracy, especially in datasets with class imbalances.
Shaeri and Katanforoush \cite{shaeri2023semi} presented a semi-supervised fake news detection method that integrates LSTM with self-attention layers. They introduced a pseudo-labeling algorithm to mitigate data scarcity, refining labels iteratively to enhance model performance. Additionally, transfer learning from sentiment analysis using pre-trained RoBERTa models further boosts accuracy. Overall, their approach demonstrates effectiveness in addressing data limitations and utilizing advanced deep learning techniques for fake news detection.
Canh et al. \cite{canh2023fake} proposed MCFC, a fake news detection method that uses multi-view fuzzy clustering on data from various sources, extracting features like title, content, and social media engagement. By applying multi-view clustering and semi-supervised learning, the method improves accuracy by incorporating diverse perspectives. The MCFC model outperforms traditional methods but faces challenges with parameter complexity and computational efficiency, especially with multi-source data.
WeSTeC \cite{akdag2024early} automates labeling by using models trained on a small labeled dataset, applying them to unlabeled data with weak labels aggregated through strategies like majority voting. It includes a RoBERTa classifier fine-tuned for the target domain and tackles semi-supervision and domain adaptation by leveraging limited labeled data and data from different domains.

Nguyen and Do \cite{nguyen2023fake} proposed a fake news detection method for Vietnamese datasets that combines a knowledge graph with semi-supervised Graph Convolutional Networks (GCN). They used GloVe embeddings, Word Mover’s Distance (WMD), and KNN to construct the knowledge graph, which is then used with GCN for detection.
MODEL \cite{liu2024rumor} detects rumors by learning source tweet representations through user correlation and information propagation. Using Graph Neural Networks, it extracts representations from a bipartite graph of user-tweet relationships and a tree structure representing information spread. By combining these representations, MODEL achieves high accuracy in rumor classification.
Inan \cite{inan2022zoka} proposed ZoKa, a fake news detection method that first identifies potential fake users using a user graph and Graph Attention Network (GAT). It then creates a content graph from user interactions and applies Edge-weighted Graph Attention Network (EGAT) with pre-trained encoders to detect fake news.
LOSS-GAT \cite{lakzaei2024loss} is a semi-supervised one-class approach that combines GNNs with label propagation. Initially, a two-step label propagation algorithm assigns preliminary labels to news articles, categorizing them as either fake or real. The graph structure is then refined using structural augmentation techniques to enhance its expressiveness. Finally, an improved GNN incorporates randomness in node neighborhoods during aggregation to predict labels for unlabeled data.
The GBCA model \cite{zhang2024gbca} integrates Graph Convolution Networks (GCN) with BERT and a co-attention mechanism to address the limitations of previous fake news detection approaches. Unlike methods focusing solely on propagation patterns or semantics, the GBCA model combines both dimensions—propagation structure and semantic features—by dynamically weighting them. This fusion enhances the model's performance, making it more accurate and efficient in detecting fake news across multiple datasets.

Although the aforementioned methods leverage the power of GNNs for fake news detection, they, along with other similar approaches, face inherent challenges arising from the limitations of standard message passing techniques in these networks.
Standard message passing primarily aggregates information from local neighbors, which may result in models overlooking global structural patterns in the graph.
This limitation is particularly problematic in tasks like fake news detection, where relationships across distant nodes, such as long-range dependencies in social networks, can provide critical contextual signals.
Furthermore, with an increasing number of layers, GNNs often suffer from over-smoothing, where node embeddings from different classes become indistinguishable.
This phenomenon reduces the discriminative power of the model, especially in large and densely connected graphs.
While various methods have been proposed to address these challenges in standard message passing, recent research has introduced innovative approaches to enhance message passing mechanisms in GNNs.
Among these, techniques leveraging Gumbel-Softmax have gained attention due to their ability to improve the flexibility and efficiency of message propagation.

Acharya and Zhang \cite{acharya2020feature} extended the Gumbel-Softmax feature selection algorithm to Graph Neural Networks (GNNs). They proposed a method for selecting and ranking features to reduce the feature set while maintaining high classification accuracy. 
The Learning to Propagate (L2P) \cite{xiao2021learning} framework enables GNNs to learn personalized and adaptive message propagation strategies. By utilizing latent variables, the method optimizes the propagation steps for each node individually. It further employs the Variational Expectation-Maximization (VEM) approach to estimate the maximum likelihood of the GNN parameters, enhancing the flexibility and effectiveness of the propagation process.
Chaudhary and Singh \cite{chaudhary2023gumbel} proposed a Gumbel-SoftMax-based Graph Convolutional Network (GS-GCN) for identifying hidden communities in complex networks. Their model utilizes the Gumbel-SoftMax function for feature extraction and incorporates a two-layer GCN, leveraging degree and adjacency matrices to enhance community detection.
CO-GNNs \cite{DBLP:conf/icml/FinkelshteinHBC24} propose a dynamic message-passing paradigm for graph neural networks, allowing nodes to choose actions like STANDARD, LISTEN, BROADCAST, or ISOLATE instead of uniformly propagating information. This flexibility enables adaptive information flow based on node states and graph topology. The architecture integrates an action network for selecting strategies and an environment network for updating representations.
DHGAT \cite{lakzaei2025decision}, or Decision-based Heterogeneous Graph Attention Network, is a model designed for fake news detection in a semi-supervised setting. It overcomes the limitations of traditional GNNs by dynamically optimizing and selecting the most suitable neighborhood type for each node in every layer in a heterogeneous graph. The architecture comprises a decision network that identifies the optimal neighborhood type and a representation network that updates node embeddings based on this selection.

The aforementioned methods have demonstrated the effectiveness of utilizing Gumbel-Softmax in enhancing the flexibility and efficiency of message propagation in GNNs.
However, despite these advancements, they still suffer from a fundamental limitation inherent to GNNs: the restriction of a node's message propagation to its L-hop neighbors in an L-layer network.
In such networks, a node cannot access neighbors beyond L hops, which prevents the propagation of information from distant nodes.
To address this limitation, we propose NOL-GAT, a novel architecture that extends the reach of message passing by allowing each node at every layer to select the optimal neighborhood order for updating its embedding vector.
Unlike traditional methods that are restricted to layer-wise propagation within fixed L-hop neighborhoods, NOL-GAT employs a mechanism that dynamically determines the most informative hop distance for each node.
This approach ensures that distant but semantically important nodes contribute effectively to the representation learning process, while avoiding unnecessary noise from irrelevant or redundant neighbors.
In the following sections, we detail the architecture and methodology of NOL-GAT, demonstrating its ability to address the challenges of existing GNN-based methods while achieving superior performance in fake news detection.

\section{Problem definition}
\label{sec:problem_def}
Let $D = \{d_1, d_2, \dots , d_n\}$ be the news set with $n$ news items, where each news item $d_i$ contains text content $t_i$.
We formulate the problem of fake news detection as a binary classification task utilizing a semi-supervised approach.
The class labels are represented by $y \in \{0, 1\}$, where $y = 1$ indicates fake news, and $y = 0$ indicates real news.
Following the semi-supervised approach, labels for a small subset of news, denoted as $D_L \subset D,$  are observed.
Our goal is to learn the model $F$ for predicting labels for all unlabeled data $D_U = D - D_L$, utilizing the labeled news:
\begin{equation}
	\label{eq:F}
	\hat{y} = F(d_i)
\end{equation}
where $\hat{y} \in \{0,1\}$ is the predicted label, and $d_i \in D_U$ is an unlabeled example. 
We summarize frequently used notations in \autoref{tbl:notations}.
\begin{table}[!t]
	\centering
	\caption{Notations and their descriptions.\label{tbl:notations}}	
	\small
	\begin{tabular}{l l}
		\hline
		\textbf{Notation} & \textbf{Definition} \\ \hline
		
		$D = \{d_1, d_2, \dots , d_n\}$		& set of news items \\ \hline
		
		$n$		& number of news items in $D$ \\ \hline		
		
		$D_L \subset D$		& set of labeled news items \\ \hline
		
		$D_U \subset D$		& set of unlabeled news items \\ \hline
		
		$y \in \{0 , 1 \} $		& ground-truth class label \\ \hline
		
		$\hat{y} \in \{0, 1 \}$		& predicted class label \\ \hline	
		
		$X \in \Bbb{R}^{n \times l_t}$		& textual feature vectors \\ \hline
		
		$l_t$		& length of textual feature vectors \\ \hline
		
		$V_{KNN} = \{v_1 , d_2, \dots , d_n\}$ & set of vertices \\ \hline
		
		$E_{KNN} = \{(v_i, v_j) \vert v_i , v_j \in V_{KNN} \}$ & set of edges \\ \hline
		
		$G_{KNN} = (V_{KNN}, E_{KNN})$ & constructed graph employing K-nearest neighbors algorithm \\ \hline
		
		$\Gamma = \{0, 1, 2, \dots, d_g\} $ & set of neighborhood orders in $G_{KNN}$ \\ \hline
		
		$d_g$ & diameter of $G_{KNN}$ \\ \hline		
		
		$\rho^l_v \in \Bbb{R}^{\vert \Gamma \vert}$ & \makecell[l]{probability distribution over possible \\ neighborhood orders $\Gamma$ for node $v$ } \\ \hline
		
		$N_{\Phi}(v) = \{u \vert (v,u) \in E_{KNN} \}$ & $\Phi$-hop neighbors of node $v$ \\ \hline
		
		$h^l_v \in \Bbb{R}^{d_l}$ & embedding of node $v$ in layer $l$ \\ \hline
		
		$d_l$ & size of embeddings in layer $l$ \\ \hline		
	\end{tabular}
\end{table}


\section{Proposed model}
\label{sec:proposed_model}
\begin{figure}[!t]
	\centering
	\includegraphics[scale=.75]{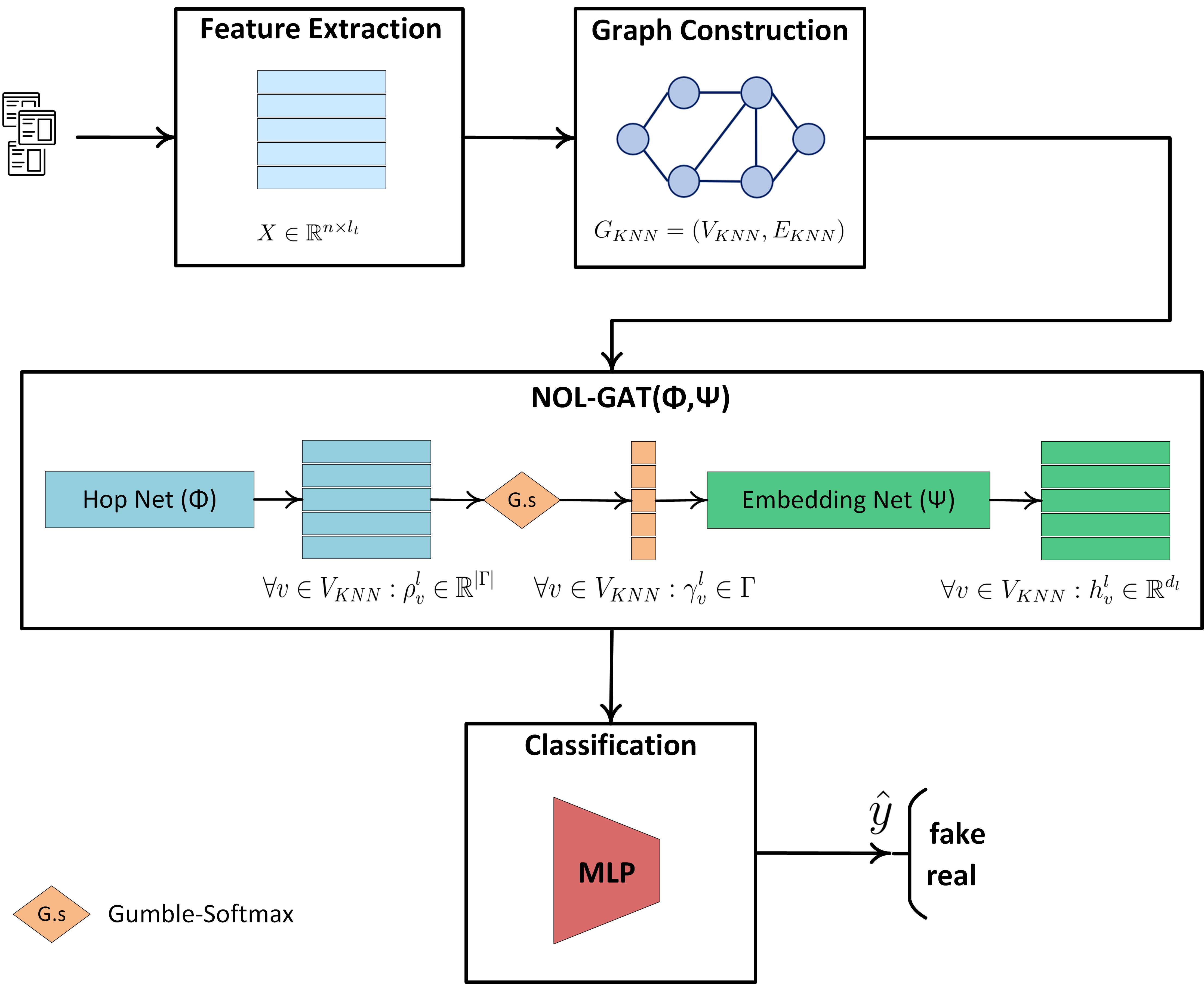}
	\caption{An overview of the proposed model NOL-GAT}
	\label{fig:dhgat}
\end{figure}

In this paper, we propose our Neighborhood-Order Learning Graph Attention Network (NOL-GAT) for detecting fake news using a semi-supervised approach.
The NOL-GAT model, illustrated in \autoref{fig:dhgat}, consists of four main modules: feature extraction, graph construction, Neighborhood-Order Learning Graph Attention Network (NOL-GAT), and classification.
Initially, the feature extraction module generates textual embeddings for the news using the Doc2Vec model.
Then, the graph construction module builds a similarity graph $G_{KNN}$ among the news items. 
In this graph, each news item is viewed as a node, and each node is connected to its $K$ nearest neighbors through undirected edges.
Then this graph is fed into the NOL-GAT model, where each node in each layer independently learns the most effective k-hop  neighborhood to update its embedding vector. 
The final embedding vectors are fed into a Multi-Layer Perceptron (MLP) classifier, to be classified as fake or real.

\subsection{Feature extraction}
\label{sec:feature_ext}
We utilize the Doc2Vec \cite{le2014distributed} approach, to analysis the textual content of news items and generate initial feature vectors for news items.
It is an embedding algorithm designed to represent entire documents or paragraphs as fixed-length numerical vectors. 
It extends Word2Vec \cite{DBLP:journals/corr/abs-1301-3781} by adding a unique identifier for each document, enabling it to capture the semantic meaning of the entire text:
\begin{equation}
	Doc2Vec(t_i) = x_i 
\end{equation}
where $t_i$ denotes textual content for news item $d_i \in D$, $x_i \in \mathbb{R}^{l_t}$ denotes the text embedding vector for $d_i$, and $l_t$ denotes the length of text’s embedding vectors.

\subsection{Graph construction}
\label{sec:graph_const}
We model $D$ (news items dataset) as a graph $G_{KNN} = (V_{KNN}, E_{KNN})$ employing K-nearest neighbors algorithm.
We view each news items $d_i \in D$ as a node $v_i \in V_{KNN}$.
Then we compute the similarity between nodes text embeddings ($\{x_1, x_2, \dots , x_n\}$) using a similarity metric such as cosine similarity.
Then every node is linked to its $k$ nearest neighbors, building an undirected similarity graph $G_{KNN}$. 
\begin{equation}
	\label{eq:G_KNN}
	G_{KNN} = (V_{KNN}, E_{KNN}):
	\begin{cases}
		V_{KNN} = \{v_1, v_2, \dots , v_n\} \\
		\vert V_{KNN} \vert = \vert D \vert \\
		E_{KNN} = \{(v_i , v_j) \vert v_j \in N_K(v_i)\}
	\end{cases}
\end{equation}
where $N_K(v_i)$ is the set of K-nearest neighbors of $v_i$.

\subsection{Neighborhood-Order Learning Graph Attention Network (NOL-GAT)}
\label{sec:KHLGAT}
\begin{figure}[!t]
	\centering
	\includegraphics[scale=.5]{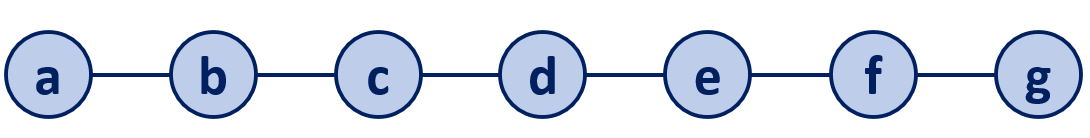}
	\caption{An illustration of the neighborhood limitation in standard GNNs. The graph consists of 7 nodes ($a$ to $g$) arranged in a simple path, where each node is connected to its adjacent neighbors. In a 2-layer GNN, node $a$ can only aggregate information from nodes $b$ and $c$, while information from more distant but potentially important nodes remains inaccessible.}
	\label{fig:sample}
\end{figure}
Graph neural networks update the target node's features through a messaging mechanism and information aggregation from its neighbors.
However, the standard GNN architecture faces challenges that can limit its performance.
One of the main challenges is the restricted access to information from more distant neighbors in the graph.
In the standard setup, a GNN with $L$ layers can at most obtain information from neighbors at a distance of $L$.
This means that if a high-importance node
(in terms of, for example, topology or features) is located farther than $L$ hops from the target node, its information cannot influence the target node.
For example, consider \autoref{fig:sample}.  
In this graph, the nodes are arranged in a simple path with 7 nodes ($a$ to $g$).
Each node is connected to its adjacent nodes,
i.e., its previous and next nodes if they exist.
Now, assume we have a 2-layer GNN:
\begin{itemize}
	\item In this network, each node can collect information from its 1-hop neighbors at each layer.  
	\item After the first layer, each node will have information from itself and its 1-hop neighbors.  
	\item After the second layer, each node will have information from its 1-hop and 2-hop neighbors.  
\end{itemize}
If the target node is $a$, after two layers of the GNN, this node will only be able to access information from nodes $b$ and $c$.
The information from other nodes will remain inaccessible to node $a$, even if these nodes hold important information.
Increasing the number of layers in a GNN is proposed as a naive solution to this issue, but this approach comes with serious problems:
\begin{itemize}
	\item \textbf{Over-squashing:}
	As the number of layers increases, information collected from a large number of nodes is compressed into embeddings of fixed and limited size.
	This compression leads to the loss of key details in the information \cite{DBLP:conf/iclr/0002Y21}.
	
	\item \textbf{Over-smoothing:}
	In deep graph networks, as the number of layers increases, the embeddings of nodes become increasingly similar, reducing their separability \cite{li2018deeper}.
	
	\item \textbf{Computational complexity:}
	Increasing the depth of the network leads to higher computational costs and reduces the model's efficiency.
\end{itemize}
To address these challenges, we introduce a new and adaptive architecture for GNNs called NOL-GAT
(Neighborhood-Order Learning Graph Attention Network), which allows each node to independently learn its optimal neighborhood order at each layer, independent of other nodes and layers.
For example, suppose in \autoref{fig:sample}, the best neighbors for node $a$ are nodes $d$ and $e$, which are its 3-hop and 4-hop neighbors, respectively, and for node $g$, the best neighbors are nodes $f$ and $b$, which are its 1-hop and 5-hop neighbors, respectively.
Therefore, the optimal embedding vectors for these two nodes are generated when these specific neighborhoods are used to update the embedding vectors of these nodes.
In a standard 2-layer GNN, such an option is not available.
However, NOL-GAT, with its capability to learn the optimal neighborhood order at each layer for each node, enables this functionality.

In NOL-GAT, we need to predict categorical neighborhood orders for each node at each layer.
Therefore, each node at each layer can choose k-hop neighbors such that $k \in \Gamma = \{0, 1, 2, \dots, d_g\}$, where $d_g$ is the diameter of the graph (the length of the longest shortest path between any two nodes in the graph).
However, it is clear that not all nodes necessarily have a $d_g$-hop neighborhood, and the maximum neighborhood order varies across different nodes.
Nodes located at the center or in denser sections of the graph typically have access to all other nodes within a distance shorter than the graph's diameter.
Only some nodes, usually those at the ends of the graph or closer to the structural boundaries of the graph, may have a $d_g$-hop neighborhood.
For example, in the graph shown in \autoref{fig:sample}:
\begin{itemize}
	\item Node $a$ and node $g$, which are at the ends of the graph, specifically have a 6-hop neighborhood, since these two nodes are at a distance equal to the diameter of the graph from each other.
	\item Nodes like $d$, located at the center of the graph, do not require such a high-order neighborhood, as their distance from all other nodes is less than the diameter of the graph.	
\end{itemize} 

Predicting categorical neighborhood orders for each node presents a challenge for gradient-based optimization because of its non-differentiable nature.
The Gumbel-Softmax estimator \cite{DBLP:conf/iclr/JangGP17, DBLP:conf/icml/FinkelshteinHBC24, DBLP:conf/iclr/MaddisonMT17} tackles this issue by providing a continuous and differentiable approximation of discrete sampling.
Our objective is to learn a categorical distribution over $\Gamma = \{0, 1, 2, \dots, d_g\}$, represented by a probability vector $\rho \in \mathbb{R}^{\vert \Gamma \vert}$ where each element corresponds to the probability of a specific neighborhood order $\gamma_i \in \Gamma$.
Let $\rho (\gamma_i)$ represent the probability of neighborhood order $\gamma_i \in \Gamma$, and let $e$ denote a one-hot vector representing the selected neighborhood order.
To sample $e$ from a categorical distribution, the Gumbel method offers a straightforward and effective approach:
\begin{equation}
	\label{eq:gumblel}
	e = \text{one-hot}\left(\underset{\gamma_i \in \Gamma}{\text{arg max}}\left(g(\gamma_i) + \log(\rho(\gamma_i))\right)\right).
\end{equation}

Here, $g(\gamma_i) \sim \text{Gumbel}(0, 1)$ is an i.i.d. sample for neighborhood type $\gamma_i$.
The $Gumbel(0, 1)$ distribution can be sampled through inverse transform sampling by drawing $u_i \sim uniform(0,1)$ and calculating $g(\gamma_i) = - \log(- \log(u_i))$ \cite{DBLP:conf/iclr/JangGP17}.
The Gumbel-Softmax estimator uses the softmax function as a continuous, differentiable alternative to $arg max$.
Thus, the Gumbel-Softmax scores are calculated as \cite{DBLP:conf/icml/FinkelshteinHBC24}:
\begin{equation}
	\label{eq:gumble-sm}
	\text{Gumbel-softmax} (\rho; \iota) =
	\frac{exp((log(\rho) + g)/\iota)}
	{\sum_{\gamma_i \in \Gamma} exp((log(\rho(\gamma_i)) + g(\gamma_i))/\iota)},	
\end{equation}
In this equation, $\rho \in \mathbb{R}^{|\Gamma|}$ represents the categorical distribution,
$g \in \mathbb{R}^{|\Gamma|}$ is the Gumbel-distributed vector,
and $\iota$ is the temperature parameter.
As the temperature $\iota$ decreases, the resulting vector becomes closer to a one-hot vector.
The Straight-Through Gumbel-Softmax estimator applies the Gumbel-Softmax approximation during the backward pass for differentiable updates, while it uses regular sampling during the forward pass.

Formally, the NOL-GAT($\Phi, \Psi$) architecture utilizes the combination and simultaneous training of two networks:
1) the hop network $\Phi$ which learns the optimal neighborhood order for each node, and 
2) the embedding network $\Psi$ which updates the embedding vectors of nodes based on the neighborhood determined by network  $\Phi$.  
Thus, the NOL-GAT network, for the graph $G_{KNN}$ (\autoref{eq:G_KNN}), updates the embedding vector of each node, such as $v$, in two stages as follows:
\begin{itemize}
	\item \textbf{Learning the neighborhood order with hop network $\Phi$:}	 
	In this step, the adaptive learning mechanism allows each node to select an optimal neighborhood order from its neighbors at various levels, such as 1-hop, 2-hop, 3-hop, and so on. This selection is performed using the Gumbel Softmax technique, which enables a probabilistic yet differentiable choice based on the learned probability distribution.
	
	So, the hop network $\Phi$ predicts a probability distribution $\rho^l_v \in \mathbb{R}^{|\Gamma|}$ over possible neighborhood orders $\Gamma$ for node $v$ :
	\begin{equation}
		\label{eq:p_l}
		\rho^l_v = GATv2\left(h^{l-1}_v , \{h^{l-1}_u | u \in N_{\phi}(v)\}\right),
	\end{equation}	
	where ${GATv2}$ is a GATv2 network \cite{DBLP:conf/iclr/Brody0Y22}, and $N_{\phi}(v)$ is defined as the set of all $\phi$-hop neighbors of node $v$.
	The $\phi$ value is defined by the user.
	The embedding vectors in GATv2 are computed using \cite{DBLP:conf/iclr/Brody0Y22}:
	\begin{equation}
		\label{eq:gat_1}
		z_v^l = GATv2\left(z^{l-1}_v , \{z^{l-1}_u | u \in N(v)\}\right) = \parallel_{k=1}^{K} \sigma \left(\sum_{u \in N(v)} \alpha_{vu}^k W^{l-1} z_u^{l-1} \right)
	\end{equation}
	and
	\begin{equation}
		\label{eq:gat_2}
		\alpha_{vu} = a \, \sigma \left(W^{l-1} z_v^{l-1} , W^{l-1} z_u^{l-1} \right),
	\end{equation}
	where $z_v^l$ represents the embedding of node $v$ in layer $l$, $\parallel$ denotes concatenation, $K$ is the number of attention heads, $\sigma$ is a non-linear activation function,
	$N(v)$ is the set of neighbors of $v$,
	and $\alpha_{vu}$ is the attention coefficient between $v$ and $u$.
	The weight matrix $W$ and parameter $a$ are learned during training.
	Next, a neighborhood order $\gamma^l_v$ is sampled from $\rho^l_v$ using the Straight-Through Gumbel-Softmax estimator.
	In other words, for each node $v$, a one-hot vector of length $\vert \Gamma \vert$ is generated, indicating which neighborhood order is selected for that node in the current layer.
	This process enables each node to make independent decisions about gathering information from its neighbors.	
		
	\item \textbf{Updating embeddings with embedding network $\Psi$:}
	After selecting the appropriate neighborhood order for each node ($N_{\phi}(v)$), embedding network $\Psi$ updates the embedding vectors of the nodes based on their optimal neighborhoods:
	\begin{equation}
		\label{eq:h}
		h^l_v = GATv2\left(h^{l-1}_v , \{h^{l-1}_u | u \in N_{\psi}(v)\}\right),
	\end{equation}
	
	where $N_{\psi}(v)$ is the set of all $\psi$-hop neighbors of node $v$, such that $\psi$ represents the most appropriate order of neighborhood for node $v$, determined by the hop network $\Phi$.
\end{itemize}

The training process of NOL-GAT is outlined in \autoref{alg:KHLGAT}. This algorithm takes as input the graph $G_{knn}$ and the desired neighborhood order $k$ for the hop network.  
During each training iteration, the hop network $\Phi$, which in this work is implemented as a GATv2 network, utilizes the embedding vectors generated in the previous layer for each node along with its neighbors up to the specified neighborhood order $k$.
This process results in a vector of $|\Gamma|$ dimensions for each node, representing the probability distribution over the neighborhood orders $\gamma_i \in \Gamma$.  
Next, the optimal neighborhood order for each node is determined using the Gumbel-Softmax function.
In the subsequent step, the embedding network $\Psi$, also implemented as a GATv2 network in this work, updates each node’s embedding based on its previous embedding and the optimal neighbors selected by the hop network $\Phi$.  
Throughout the training process, the hop network $\Phi$ learns to identify the most suitable neighborhood order for each node, while the embedding network $\Psi$ optimizes the weights needed to update the node embeddings according to the optimal neighborhood order.
\begin{algorithm}[bht!]
	\small
	\caption{Training process of NOL-GAT.
	\label{alg:KHLGAT}}
	\SetKwInput{KwInput}{Input}                
	\SetKwInput{KwOutput}{Output}              
	\DontPrintSemicolon
	
	\KwInput{\\		
		$G_{KNN} = (V_{KNN}, E_{KNN})$: dataset graph  \\		
		$X \in \Bbb{R}^{n \times l_t}$: text embeddings vectors \\
		$max\_epoch$ : number of epochs \\
		$L$: number of GNN layers \\
		$k \in \Gamma$: neighborhood order for hope network $\Phi$					
	}
	\BlankLine	
	\KwOutput{Trained model $NOL-GAT(\Phi, \Psi)$}
	\BlankLine
	\BlankLine
	\BlankLine
	\ForEach{$v \in V_{KNN}$}
	{	
		$N_k(v) = \{u \vert (v,u) \in E_{KNN} , \text{u is a k-hop neighbor of v} \}$ 
	}
	
	\BlankLine
	\BlankLine
	
	\tcp{Train NOL-GAT on $G_{KNN}$}
	\ForEach{$epoch : 1..max\_epoch$}
	{	\ForEach{$v \in V_{KNN}$}
		{	
			$h^0_v = X_v$
		}
		\ForEach{$l: 1..L$}
		{	\ForEach{$v \in V_{KNN}$}
			{
				\tcp{Learn the optimal neighborhood order using hop network $\Phi$}
				
				$\rho^l_v = GATv2(h^{l-1}_v , \{h^{l-1}_u | u \in N_k(v)\})$ \tcp{\autoref{eq:p_l}}	
				$\gamma^l_v = \text{Gumble-Softmax}(\rho^l_v)$	\\
				
				\BlankLine
				\BlankLine
				
				\tcp{Update embeddings using embedding network $\Psi$}
				
				$N_r(v) = \{u \vert (v,u) \in E_{KNN} , r=\gamma^l_v, \text{ u is an r-hop neighbor of v} \} $ \\
				\BlankLine
				$h^l_v = GATv2(h^{l-1}_v , \{h^{l-1}_u | u \in N_r(v)\})$ \tcp{\autoref{eq:h}}									
			}
		}
		\BlankLine
		$H^L = [h^L_1, \dots , h^L_n]^T$ \tcp{$n = |V_{KNN}|$}
		$\hat{Y} = MLP(H^L)$	\\
		$loss \leftarrow \autoref{eq:loss}$ \\
		$loss.backward()$			
	}
\end{algorithm}

\subsection{Classification}
\label{sec:cls}
To classify news, the embeddings $H^L$, produced by NOL-GAT in its final layer (Layer $L$), are passed through a Multi-Layer Perceptron (MLP) network.  
The following loss function is used to train NOL-GAT:  
\begin{equation}
	\label{eq:loss}
	L(\Theta) = -\frac{1}{|D_{L}|} \sum_{i=1}^{|D_{L}|} \big( y_i \log(\hat{y}_i) + (1 - y_i) \log(1 - \hat{y}_i) \big).
\end{equation}
where $D_{L}$ represents the labeled dataset, $y_i$ and $\hat{y_i} $ denote the true label and the predicted label for the data point $d_i \in D$, respectively, and $\Theta$ consists of the parameters learned by the model during training.

\section{Model properties}
\label{sec:model_properties}

In this section, we present the key features of the proposed model.
The NOL-GAT model offers unique characteristics that differentiate it from other GNN-based models.
These features are as follows:
\begin{enumerate}
	\item \textbf{Independence of decision-making at node and layer level} 
	
	One of the key features of the NOL-GAT model is that the selection of the neighborhood order for each node is made independently of other nodes in the same layer. This means that nodes can select their neighbors based on their own specific characteristics without being influenced by other nodes. These decisions are made independently for each layer. For example, a node in the first layer may use neighbors up to a 4-hop distance, while in the second layer, the same node may use closer neighbors (e.g., 2-hop). This feature allows NOL-GAT to make more optimal decisions at each layer and adapt to different graph conditions.
	
	\item \textbf{Flexibility in neighbor selection}
		
	NOL-GAT uses an adaptive approach that allows each node to select the optimal neighbors based on its position in the graph. This adaptive approach is particularly beneficial in unbalanced graphs or graphs with nodes of varying degrees. As a result, the model can gather more precise information from its neighbors and prevent the transfer of irrelevant information.
	
	\item \textbf{Reduction of over-squashing and improved information flow}	
	
	One of the main challenges in GNNs is the issue of over-squashing, which occurs when important information from distant nodes is excessively compressed and fails to be transferred correctly to the target node. In NOL-GAT, the targeted and limited selection of neighbors helps mitigate this issue, as information from distant nodes is transferred to the target node without excessive squashing. This approach ensures that information from distant nodes reaches the target nodes more easily, without the need for complex routing or communication in deep graphs.
	
	\item \textbf{Reduction of over-smoothing}	
	
	In GNNs, over-smoothing occurs when the features of nodes become overly similar as they propagate through multiple layers, causing semantic distinctions between nodes to vanish. This phenomenon typically arises when the number of layers increases, and messages from nodes are spread horizontally and vertically across the graph. In NOL-GAT, the optimal neighborhood order selection for each node in every layer helps mitigate over-smoothing. This is because the model allows each node to select neighbors based on its own needs and position, and this targeted selection prevents the features of nodes from becoming too homogeneous across layers. This approach helps maintain feature diversity and prevents over-similarity between features as they propagate through layers, allowing nodes to retain their distinctive characteristics during the update process.
	
	\item \textbf{Reduction of computational complexity}
		
	Unlike traditional models that require increasing the number of layers to achieve high accuracy or increase computational complexity by transferring information through long paths, NOL-GAT keeps computational complexity under control by using a targeted selection of neighbors at each layer. This reduction in complexity allows the model to perform better without the need for excessive layers or network complexity.
\end{enumerate}

\section{Experiments}
\label{sec:exp}

In this section, we first present an overview of the datasets, baseline models, and evaluation metrics employed in our empirical analysis.
Next, we provide a detailed report on the results of our extensive experiments, conducted under various settings and conditions to thoroughly evaluate the effectiveness of our approach \footnote{The code of our method is publicly available at: \url{https://github.com/blakzaei/NOL-GAT}.}.

\subsection{Datasets}
\label{sec:ds}
We employ five fully labeled datasets to validate our proposed approach: two in Portuguese and three in English:
\begin{itemize}
	\item \textbf{Fake.Br\footnote{\url{https://github.com/roneysco/Fake.BR-Corpus}}} is a Portuguese dataset created for fake news detection, consisting of 7,200 articles evenly divided into 3,600 fake and 3,600 real news.
	These articles span six categories: politics (58\%), TV and celebrities (21.4\%), society and daily life (17.7\%), science and technology (1.5\%), economy (0.7\%), and religion (0.7\%) \cite{silva2020towards}.
	The dataset is available in both complete and truncated formats, and we utilize the truncated version to ensure consistent text length and avoid potential biases during learning.	
	
	\item \textbf{Fact-checked News\footnote{\url{https://github.com/GoloMarcos/FKTC}}} is a collection of 2,168 Brazilian political news articles curated between August 2018 and May 2019, a period marked by heightened disinformation around Brazil’s presidential elections.
	This dataset, sourced from platforms such as AosFatos, Agência Lupa, and UOL Confere, includes 1,124 real and 1,044 fake news items. 
	 
	\item \textbf{FakeNewsNet (FNN)} is derived from the FakeNewsNet repository\footnote{\url{https://github.com/KaiDMML/FakeNewsNet}} \cite{shu2020fakenewsnet} and comprises 21,032 English news articles fact-checked by the GossipCop\footnote{\url{https://www.gossipcop.com}} website.
	However, due to a significant token imbalance within the articles, we filter the dataset to include articles containing 200–600 tokens after removing extraneous characters and stopwords.
	This refined version consists of 1,705 fake news articles and 5,298 real news articles, reflecting a pronounced class imbalance.  
	
	\item \textbf{FakeNewsDetection\footnote{\url{https://www.kaggle.com/jruvika/fake-news-detection}}} is a Kaggle dataset uploaded by Jruvika, initially containing 4,009 news items.
	After cleaning entries with missing values, 3,988 articles remain, including 2,121 fake and 1,867 real news items.  
	
	\item \textbf{FakeNewsData\footnote{\url{https://www.kaggle.com/c/fake-news/data}}} was originally released on Kaggle in 2018 as part of a competition and includes 20,800 samples.
	After eliminating incomplete entries, the dataset contains 20,700 news articles with an almost balanced distribution of 10,360 fake and 10,340 real entries.  
\end{itemize}

To prepare the datasets for analysis, we perform standard text preprocessing steps, such as converting text to lowercase, removing stopwords, links, and numerical content, and applying stemming with the PorterStemmer from NLTK.
Finally, each article is transformed into a 500-dimensional feature vector using the Doc2Vec technique.
\autoref{tbl:ds} provides a summary of the datasets' characteristics.
\begin{table}[!t]
	\centering
	\small
	\caption{Summary of the news datasets used in the experimental evaluation.}
	\label{tbl:ds}
	\begin{tabular}{l l l l l}
		\hline
		\textbf{Dataset} & \textbf{Language} & \textbf{Total Entries} & \textbf{Fake News} & \textbf{Real News} \\ \hline		
		Fake.Br             & Portuguese & 7200  & 3600  & 3600  \\ \hline
		Fact-checked News   & Portuguese & 2168  & 1044  & 1124  \\ \hline
		FakeNewsNet (FNN)         & English    & 7003  & 1705  & 5298  \\ \hline
		FakeNewsDetection & English    & 3988  & 2121  & 1867  \\ \hline
		FakeNewsData      & English    & 20700 & 10360 & 10340 \\ \hline		
	\end{tabular}
\end{table}
\subsection{Baseline models}
\label{sec:baselines}

We assess the performance of our proposed model by comparing it against six semi-supervised fake news detection methods:
\begin{itemize}
	\item \textbf{CO-GNN \cite{DBLP:conf/icml/FinkelshteinHBC24}} is a graph neural network framework where each node dynamically chooses one of four actions (Listen, Broadcast, Standard, or Isolate) at each layer to control the flow of information.
	The nodes update their states based on their own actions and the actions of neighboring nodes, enabling more flexible and adaptive message-passing compared to traditional methods.
	\item \textbf{L2Q \cite{xiao2021learning}} models the probability of quitting at each propagation step using a stick-breaking process, assigning a non-parametric probability to each step.
	This approach helps prevent the use of deeper layers and allows the model to automatically select the optimal propagation strategy for each node.
	\item \textbf{TGNcl \cite{li2024graph}}  is a text classification model that transforms each text into a heterogeneous graph of words and labels, called a WL-graph.
	It learns text representations by encoding the graph using a message passing process, and then applies contrastive learning to enhance the feature differentiation.
	The model also employs graph augmentation techniques such as node drop, edge drop, and shuffle to improve robustness and accuracy.
	\item \textbf{LSTM-CP-GCN \cite{ren2023fake}} investigates intra-article interactions by modeling each article as a weighted graph, where sentences serve as vertices.
	An LSTM generates feature vectors for the vertices, while a CP decomposition-based approach computes the weight matrix using local word co-occurrence information between sentences.
	The resulting graph is then input into a GCN for classification.
	\item \textbf{TextGCN \cite{yao2019graph}} models the entire corpus as a heterogeneous graph that includes both word and news nodes.
	It employs a GCN to learn embeddings for both node types, effectively capturing the relationships between words and news items to perform news classification.
	\item \textbf{GATv2 \cite{DBLP:conf/iclr/Brody0Y22}} utilizes the standard message passing mechanism in GATv2 to learn embeddings, ultimately applying them for news classification.
\end{itemize}

\subsection{Experimental setup}
\label{sec:setup}
For feature extraction from news text content, we utilize the Doc2Vec model, generating 500-dimensional vectors.
The graph is constructed using the KNN method with seven different values for K (3, 4, 5, 6, 7,8).
A semi-supervised learning strategy is adopted, where three scenarios with different proportions of labeled data (10\%, 20\%, and 30\%) are considered.
As mentioned in \autoref{sec:KHLGAT} The proposed model consists of two networks,
hop network ($\Phi$) and embedding network ($\Psi$), both implemented as a two-layer GATv2 with 128 and 64 hidden units, respectively.
The model is trained using the Adam optimizer with a weight decay strategy over 200 epochs.

To assess model performance, macro-F1 and Accuracy metrics are used.
Each model (NOL-GAT and all Baselines) is executed 10 times, and the average results are reported.
The accuracy metric is defined as follows:
\begin{equation}
	\label{eq:acc}
	Acc = \frac{(TP+TN)}{(TP+FP+FN+TN)}.
\end{equation}
Here, $TP$, $TN$, $FP$ and $FN$ indicate True Positive, True Negative, False Positive, and False Negative, respectively.
In scenarios where data distribution across different classes is imbalanced, relying solely on accuracy for evaluation becomes unwise.
This is because methods that predominantly classify new samples into the majority class may achieve high accuracy while failing to properly distinguish the minority class \cite{wu2020rumor}.
Since fake news detection often involves imbalanced datasets, including the FNN dataset used in our study, we adopt macro-F1 as another evaluation metric.
The macro-F1 score for a dataset with k classes is computed through the following equations:
\begin{equation}
	P_i = \frac{TP_i}{TP_i + FP_i}, \, \, \, R_i = \frac{TP_i}{TP_i + FN_i}
\end{equation}

\begin{equation}
	P_{Macro} = \frac{\sum_{i=1}^{k} P_i}{k} ,
	\, \, \,
	R_{Macro} = \frac{\sum_{i=1}^{k} R_i}{k}
\end{equation}
\begin{equation}
	Macro-F1 = \frac{2 P_{Macro} \times R_{Macro}}{P_{Macro} + R_{Macro}}
\end{equation}

The intereset-F1  is defined as follows:
\begin{equation}
	P_{Interest} = \frac{TP_{Interest}}{TP_{Interest} + FP_{Interest}}	
\end{equation}

\begin{equation}
	R_{Interest} = \frac{TP_{Interest}}{TP_{Interest} + FN_{Interest}}
\end{equation}

\begin{equation}
	Interest\text{-}F1 = \frac{2 \times P_{Interest} \times R_{Interest}}{P_{Interest} + R_{Interest}}
\end{equation}

All methods, including the proposed approach and baseline models, are implemented in Python using the PyTorch library.
For a fair comparison, the key hyperparameters of baseline methods are set to match those of the proposed method (NOL-GAT).
GATv2 follows an identical architecture to NOL-GAT, except that it employs the standard message-passing mechanism.
Similarly, for other GNN-based approaches such as Co-GNN, L2Q, and TGNCL, the same architecture and hyperparameter settings are applied, differing only in their message-passing strategies, as discussed in \autoref{sec:baselines}.
For TextGCN and LSTM-CP-GCN, a two-layer GCN with the same number of hidden units is used, following the original model architecture.
All other configurations, including the number of epochs, learning rate, and execution repetitions, remain identical across all methods.
\subsection{Results}
\label{sec:results}
The performance of the NOL-GAT model is evaluated against six baseline models across five benchmark datasets: Fake.Br, Fact-checked News, FNN, FakeNewsDetection and FakeNewsData.
The evaluation metrics considered are accuracy and macro-F1 score, with labeled data proportions of 10\%, 20\%, and 30\%.

On the FakeBr dataset, as can be seen in \autoref{tab:results_fakebr},
NOL-GAT significantly outperforms all the competing models across all the label proportions.
With 10\% labeled data, NOL-GAT achieves an accuracy of 0.8126 and a macro-F1 score of 0.8139, whereas the closest competitor, TextGCN, obtains an accuracy of 0.6235 and a macro-F1 score of 0.6112.
As the label proportion increases to 30\%, NOL-GAT further improves its performance, achieving an accuracy of 0.8287 and a macro-F1 score of 0.8327.
This demonstrates the robustness of NOL-GAT in handling fake news detection in low-resource settings, maintaining a substantial margin over other methods, including Co-GNN and GATv2.

On the Fact-checked News dataset, as seen in \autoref{tab:results_factchecked}, NOL-GAT again exhibits superior performance, with its accuracy increasing from 0.9358 (10\% labeled data) to 0.9501 (30\% labeled data).
The macro-F1 scores follow a similar trend, with NOL-GAT achieving 0.9382 at 10\% labeled data and 0.9513 at 30\%.
Co-GNN, the second-best performer, lagged behind with an accuracy of 0.9096 (10\%) and 0.9282 (30\%). 
These results highlight NOL-GAT's capability to capture intricate patterns in graph-based representations of fake news, outperforming traditional graph neural networks like TextGCN and GATv2 by a notable margin.

On the FNN dataset, as seen in \autoref{tab:results_fnn}, NOL-GAT continues to demonstrate state-of-the-art performance.
At 10\% labeled data, it achieves an accuracy of 0.8525 and a macro-F1 score of 0.7772, outperforming Co-GNN, which obtains 0.8139 and 0.6662, respectively.
With 30\% labeled data, NOL-GAT further improves to 0.8589 accuracy and 0.7916 macro-F1.
The results indicate that NOL-GAT effectively generalizes across different datasets and maintains consistent improvements over existing approaches, particularly in settings with varying label availability.

For the FakeNewsDetection dataset, as seen in \autoref{tab:results_fnd},
NOL-GAT exhibits strong and consistent results.
It attains an accuracy of 0.8892 and a macro-F1 score of 0.8908 at 10\% labeled data, increasing to 0.9014 accuracy and 0.904 macro-F1 at 30\%.
Compared to Co-GNN, which scores 0.8 and 0.7953 at 10\%, and 0.83 and 0.8264 at 30\%, NOL-GAT consistently maintains a competitive edge.
The relatively lower performance of other models such as L2Q, TextGCN, and LSTM-CP-GCN further underscores the effectiveness of NOL-GAT's neighborhood selection mechanism.

Finally, on the FakeNewsData dataset, as seen in \autoref{tab:results_fndata},
NOL-GAT demonstrates its ability to achieve high performance across different proportions of labeled data.
With 10\% labeled data, it attains an accuracy of 0.8837 and a macro-F1 score of 0.8837, surpassing all the baseline models.
As the labeled data increases to 30\%, its accuracy improves to 0.8947, with a macro-F1 score of 0.8957.
Co-GNN and GATv2, the next best-performing models, show  lower overall performance, indicating that NOL-GAT's unique ability to adaptively learn the best neighborhood for each node, contributes to superior classification results.

In summary, across all the datasets, NOL-GAT consistently outperforms
all the baseline models, demonstrating its robustness and effectiveness in fake news detection. Its ability to dynamically learn the optimal k-hop neighborhood contributes to its strong results, particularly in datasets with varying proportions of labeled data. These findings validate NOL-GAT's potential as an advanced solution for fake news detection in graph-based settings, offering significant improvements over existing approaches.

\begin{table}[]
	\centering
	\caption{Performance comparison over Fake.Br}
	\label{tab:results_fakebr}
	\resizebox{\textwidth}{!}{%
		\begin{tabular}{ccccccc}
			\hline
			\multirow{2}{*}{method} & \multicolumn{2}{c}{10\%}      & \multicolumn{2}{c}{20\%}      & \multicolumn{2}{c}{30\%}      \\ \cline{2-7} 
			& ACC           & Macro-f1      & ACC           & Macro-f1      & ACC           & Macro-f1      \\ \hline
			NOL-GAT                  & 0.8126$\pm$0.0051 & 0.8139$\pm$0.0037 & 0.8224$\pm$0.0041 & 0.8275$\pm$0.0031 & 0.8287$\pm$0.0038 & 0.8327$\pm$0.0034 \\ \hline
			Co-GNN & 0.6166$\pm$0.0233 & 0.6116$\pm$0.0263 & 0.6425$\pm$0.0156 & 0.6394$\pm$0.0139 & 0.654$\pm$0.0179  & 0.6515$\pm$0.0185 \\ \hline
			L2Q    & 0.5701$\pm$0.0235 & 0.5636$\pm$0.0266 & 0.5837$\pm$0.0224 & 0.5782$\pm$0.0248 & 0.6026$\pm$0.0244 & 0.5994$\pm$0.0261 \\ \hline
			TextGCN                 & 0.6235$\pm$0.0315 & 0.6112$\pm$0.0424 & 0.6616$\pm$0.0198 & 0.6542$\pm$0.0225 & 0.7018$\pm$0.0116 & 0.6996$\pm$0.0125 \\ \hline
			TGNCL  & 0.5673$\pm$0.0293 & 0.5581$\pm$0.033  & 0.53$\pm$0.0394   & 0.4364$\pm$0.0866 & 0.4964$\pm$0.0125 & 0.3372$\pm$0.0089 \\ \hline
			LSTM-CP-GCN             & 0.5153$\pm$0.0138 & 0.511$\pm$0.0162  & 0.5247$\pm$0.0134 & 0.522$\pm$0.0135  & 0.5205$\pm$0.0167 & 0.5178$\pm$0.0177 \\ \hline
			GATv2  & 0.6171$\pm$0.0179 & 0.613$\pm$0.0172  & 0.6334$\pm$0.0101 & 0.6306$\pm$0.0112 & 0.6504$\pm$0.0172 & 0.6483$\pm$0.0176 \\ \hline
		\end{tabular}%
	}
\end{table}
\begin{table}[]
	\centering
	\caption{Performance comparison over Fact-checked News.}
	\label{tab:results_factchecked}
	\resizebox{\textwidth}{!}{%
		\begin{tabular}{lllllll}
			\hline
			\multicolumn{1}{c}{\multirow{2}{*}{method}} &
			\multicolumn{2}{c}{10\%} &
			\multicolumn{2}{c}{20\%} &
			\multicolumn{2}{c}{30\%} \\ \cline{2-7} 
			\multicolumn{1}{c}{} &
			\multicolumn{1}{c}{ACC} &
			\multicolumn{1}{c}{Macro-f1} &
			\multicolumn{1}{c}{ACC} &
			\multicolumn{1}{c}{Macro-f1} &
			\multicolumn{1}{c}{ACC} &
			\multicolumn{1}{c}{Macro-f1} \\ \hline
			NOL-GAT      & 0.9358$\pm$0.0029 & 0.9382$\pm$0.0027 & 0.9458$\pm$0.0039 & 0.9468$\pm$0.0052 & 0.9501$\pm$0.005  & 0.9513$\pm$0.0036 \\ \hline
			Co-GNN      & 0.9096$\pm$0.0106 & 0.9096$\pm$0.0107 & 0.9186$\pm$0.0056 & 0.9186$\pm$0.0055 & 0.9282$\pm$0.0046 & 0.9282$\pm$0.0047 \\ \hline
			L2Q         & 0.8991$\pm$0.0122 & 0.8991$\pm$0.0123 & 0.9103$\pm$0.0118 & 0.9102$\pm$0.0118 & 0.9235$\pm$0.0067 & 0.9235$\pm$0.0067 \\ \hline
			TextGCN     & 0.8784$\pm$0.0118 & 0.8783$\pm$0.0124 & 0.8952$\pm$0.0117 & 0.8983$\pm$0.0123 & 0.9014$\pm$0.0073 & 0.9027$\pm$0.0068 \\ \hline
			TGNCL       & 0.7986$\pm$0.0373 & 0.7969$\pm$0.0379 & 0.8146$\pm$0.0432 & 0.8135$\pm$0.0438 & 0.7685$\pm$0.087  & 0.7541$\pm$0.113  \\ \hline
			LSTM-CP-GCN & 0.7628$\pm$0.0386 & 0.7616$\pm$0.0378 & 0.7799$\pm$0.0439 & 0.7767$\pm$0.0447 & 0.7336$\pm$0.0899 & 0.7174$\pm$0.1126 \\ \hline
			GATv2       & 0.8467$\pm$0.0066 & 0.8509$\pm$0.0056 & 0.856$\pm$0.0108  & 0.8628$\pm$0.0074 & 0.8641$\pm$0.0081 & 0.8661$\pm$0.0068 \\ \hline
		\end{tabular}%
	}
\end{table}
\begin{table}[]
	\centering
	\caption{Performance comparison over FakeNewsNet (FNN).}
	\label{tab:results_fnn}
	\resizebox{\textwidth}{!}{%
		\begin{tabular}{lllllll}
			\hline
			\multicolumn{1}{c}{\multirow{2}{*}{method}} &
			\multicolumn{2}{c}{10\%} &
			\multicolumn{2}{c}{20\%} &
			\multicolumn{2}{c}{30\%} \\ \cline{2-7} 
			\multicolumn{1}{c}{} &
			\multicolumn{1}{c}{ACC} &
			\multicolumn{1}{c}{Macro-f1} &
			\multicolumn{1}{c}{ACC} &
			\multicolumn{1}{c}{Macro-f1} &
			\multicolumn{1}{c}{ACC} &
			\multicolumn{1}{c}{Macro-f1} \\ \hline
			NOL-GAT      & 0.8525$\pm$0.0037 & 0.7772$\pm$0.0044 & 0.8545$\pm$0.0061 & 0.7855$\pm$0.0039 & 0.8589$\pm$0.0052 & 0.7916$\pm$0.0043 \\ \hline
			Co-GNN      & 0.8139$\pm$0.0068 & 0.6662$\pm$0.0275 & 0.824$\pm$0.0079  & 0.6727$\pm$0.0172 & 0.8234$\pm$0.0078 & 0.6788$\pm$0.0204 \\ \hline
			L2Q         & 0.806$\pm$0.0132  & 0.6622$\pm$0.0255 & 0.7952$\pm$0.0139 & 0.6456$\pm$0.0833 & 0.8001$\pm$0.0171 & 0.6828$\pm$0.0164 \\ \hline
			TextGCN     & 0.7734$\pm$0.0123 & 0.5519$\pm$0.0562 & 0.7832$\pm$0.0094 & 0.5766$\pm$0.0264 & 0.7902$\pm$0.0143 & 0.6103$\pm$0.029  \\ \hline
			TGNCL       & 0.7116$\pm$0.1152 & 0.5415$\pm$0.0612 & 0.7239$\pm$0.0773 & 0.5545$\pm$0.0627 & 0.7337$\pm$0.0771 & 0.5156$\pm$0.0618 \\ \hline
			LSTM-CP-GCN & 0.6701$\pm$0.0358 & 0.4992$\pm$0.0243 & 0.6874$\pm$0.0284 & 0.5053$\pm$0.0196 & 0.6835$\pm$0.0212 & 0.4956$\pm$0.0231 \\ \hline
			GATv2       & 0.7686$\pm$0.0088 & 0.6833$\pm$0.0083 & 0.7696$\pm$0.0048 & 0.6974$\pm$0.0114 & 0.7725$\pm$0.0049 & 0.701$\pm$0.0057  \\ \hline
		\end{tabular}%
	}
\end{table}
\begin{table}[]
	\centering
	\caption{Performance comparison over FakeNewsDetection.}
	\label{tab:results_fnd}
	\resizebox{\textwidth}{!}{%
		\begin{tabular}{lllllll}
			\hline
			\multicolumn{1}{c}{\multirow{2}{*}{method}} &
			\multicolumn{2}{c}{10\%} &
			\multicolumn{2}{c}{20\%} &
			\multicolumn{2}{c}{30\%} \\ \cline{2-7} 
			\multicolumn{1}{c}{} &
			\multicolumn{1}{c}{ACC} &
			\multicolumn{1}{c}{Macro-f1} &
			\multicolumn{1}{c}{ACC} &
			\multicolumn{1}{c}{Macro-f1} &
			\multicolumn{1}{c}{ACC} &
			\multicolumn{1}{c}{Macro-f1} \\ \hline
			NOL-GAT      & 0.8892$\pm$0.005  & 0.8908$\pm$0.0034 & 0.8999$\pm$0.0056 & 0.8994$\pm$0.0031 & 0.9014$\pm$0.0034 & 0.904$\pm$0.0029  \\ \hline
			Co-GNN      & 0.8$\pm$0.0147    & 0.7953$\pm$0.0158 & 0.819$\pm$0.012   & 0.8156$\pm$0.0121 & 0.83$\pm$0.0122   & 0.8264$\pm$0.0123 \\ \hline
			L2Q         & 0.7711$\pm$0.0262 & 0.7671$\pm$0.0263 & 0.7906$\pm$0.0193 & 0.7883$\pm$0.019  & 0.8054$\pm$0.019  & 0.8011$\pm$0.0224 \\ \hline
			TextGCN     & 0.7054$\pm$0.027  & 0.7008$\pm$0.0304 & 0.7464$\pm$0.0168 & 0.7445$\pm$0.0175 & 0.7462$\pm$0.0296 & 0.7479$\pm$0.0127 \\ \hline
			TGNCL       & 0.7545$\pm$0.0709 & 0.7362$\pm$0.1074 & 0.79$\pm$0.0375   & 0.7806$\pm$0.0462 & 0.7072$\pm$0.0815 & 0.6617$\pm$0.138  \\ \hline
			LSTM-CP-GCN & 0.7324$\pm$0.0704 & 0.712$\pm$0.1065  & 0.7659$\pm$0.0379 & 0.7569$\pm$0.0463 & 0.6836$\pm$0.0809 & 0.6383$\pm$0.1387 \\ \hline
			GATv2       & 0.801$\pm$0.01    & 0.7967$\pm$0.0112 & 0.8222$\pm$0.0075 & 0.8197$\pm$0.0076 & 0.836$\pm$0.0127  & 0.8329$\pm$0.0128 \\ \hline
		\end{tabular}%
	}
\end{table}
\begin{table}[]
	\centering
	\caption{Performance comparison over FakeNewsData.}
	\label{tab:results_fndata}
	\resizebox{\textwidth}{!}{%
		\begin{tabular}{lllllll}
			\hline
			\multicolumn{1}{c}{\multirow{2}{*}{method}} &
			\multicolumn{2}{c}{10\%} &
			\multicolumn{2}{c}{20\%} &
			\multicolumn{2}{c}{30\%} \\ \cline{2-7} 
			\multicolumn{1}{c}{} &
			\multicolumn{1}{c}{ACC} &
			\multicolumn{1}{c}{Macro-f1} &
			\multicolumn{1}{c}{ACC} &
			\multicolumn{1}{c}{Macro-f1} &
			\multicolumn{1}{c}{ACC} &
			\multicolumn{1}{c}{Macro-f1} \\ \hline
			NOL-GAT      & 0.8837$\pm$0.0042 & 0.8837$\pm$0.0058 & 0.887$\pm$0.005   & 0.8893$\pm$0.0053 & 0.8947$\pm$0.0049 & 0.8957$\pm$0.0039 \\ \hline
			Co-GNN      & 0.8113$\pm$0.0124 & 0.8106$\pm$0.0122 & 0.8159$\pm$0.0114 & 0.8153$\pm$0.0112 & 0.821$\pm$0.0078  & 0.8206$\pm$0.008  \\ \hline
			L2Q         & 0.7747$\pm$0.0236 & 0.7739$\pm$0.0238 & 0.7905$\pm$0.0185 & 0.7893$\pm$0.0191 & 0.7979$\pm$0.0136 & 0.7975$\pm$0.0138 \\ \hline
			TextGCN     & 0.7362$\pm$0.0248 & 0.7365$\pm$0.0249 & 0.7498$\pm$0.0203 & 0.7507$\pm$0.0179 & 0.757$\pm$0.0128  & 0.7582$\pm$0.0143 \\ \hline
			TGNCL       & 0.7347$\pm$0.0243 & 0.7365$\pm$0.0238 & 0.7513$\pm$0.0182 & 0.7499$\pm$0.021  & 0.758$\pm$0.0156  & 0.7577$\pm$0.0138 \\ \hline
			LSTM-CP-GCN & 0.7035$\pm$0.0263 & 0.7041$\pm$0.0227 & 0.7156$\pm$0.0183 & 0.7178$\pm$0.0206 & 0.7216$\pm$0.0146 & 0.7239$\pm$0.0129 \\ \hline
			GATv2       & 0.8059$\pm$0.0109 & 0.8049$\pm$0.0113 & 0.8103$\pm$0.0098 & 0.8098$\pm$0.0097 & 0.8212$\pm$0.0053 & 0.8206$\pm$0.0053 \\ \hline
		\end{tabular}%
	}
\end{table}
\subsubsection{Analysis of the impact of adaptive message passing in NOL-GAT:
	a comparative study with standard GATv2}
\label{sec:khlgat-vs-gatv2}
\begin{figure}[!t]
	\centering
	\subfigure[]{\includegraphics[width=0.90\textwidth]{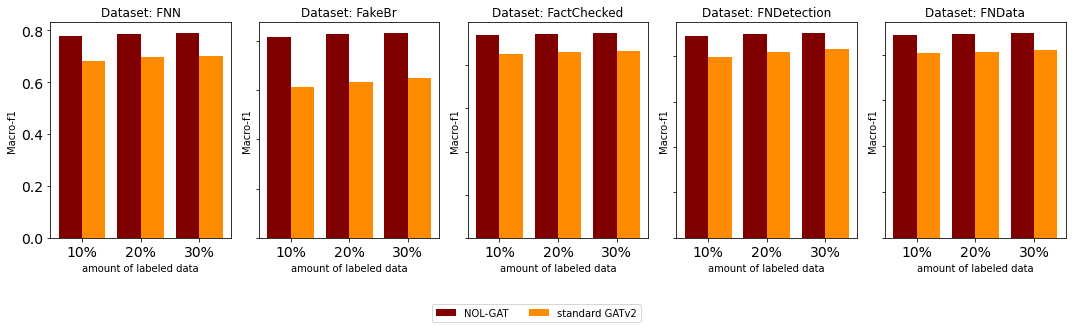}}
	\subfigure[]{\includegraphics[width=0.90\textwidth]{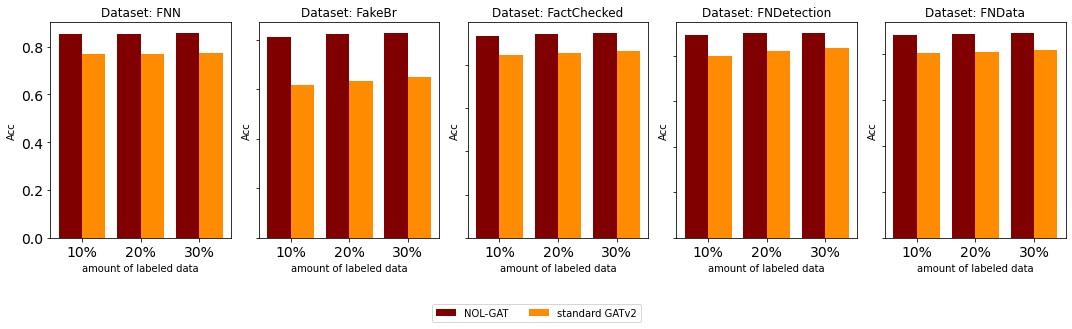}}
	\caption{Macro-F1 (a) and accuracy (b) comparision of NOL-GAT and standard GATv2.}
	\label{fig:khlgta-vs-gatv2}
\end{figure}
\autoref{fig:khlgta-vs-gatv2} presents the comparison results between the proposed NOL-GAT model and the baseline GATv2 model in terms of accuracy and macro-F1.
This comparison, conducted on five different datasets, demonstrates that NOL-GAT consistently outperforms GATv2 across all the cases.
To ensure a fair comparison, both models share the same overall architecture, meaning that the number of layers, neurons, and other settings are identical, with the only difference being the message-passing mechanism.
In NOL-GAT, each node independently learns the optimal k-hop neighborhood, whereas GATv2 employs the standard message-passing approach.
This modification enhances the learning process, leading to improved accuracy for the proposed model.
Furthermore, the evaluation follows a semi-supervised setting, where three different levels of labeled data (10\%, 20\%, and 30\%) are considered.
The results indicate that NOL-GAT consistently outperforms GATv2 at all levels of labeled data.
Additionally, while both models benefit from an increase in labeled data, the superiority of NOL-GAT remains evident in all the scenarios.
This trend suggests that the proposed model is better at leveraging the graph structure and integrating information from both labeled and unlabeled data, ultimately achieving higher generalization capability compared to standard GATv2.

\subsubsection{Analysis of the impact of the quantity of labeled data}
\label{sec:labeled}
\begin{figure}[!t]
	\centering
	\subfigure[]{\includegraphics[width=0.48\textwidth]{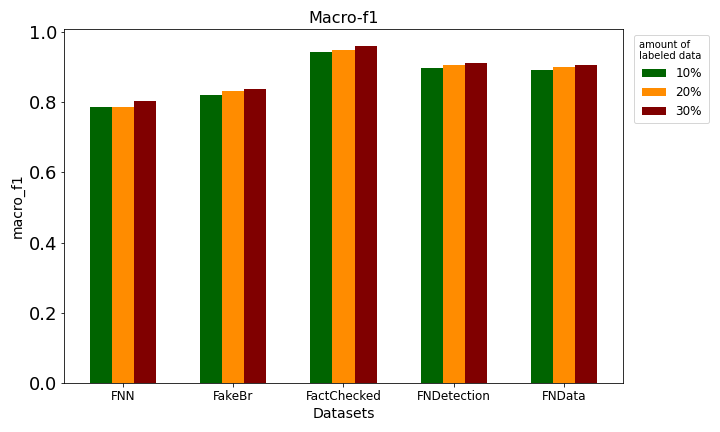}}
	\subfigure[]{\includegraphics[width=0.48\textwidth]{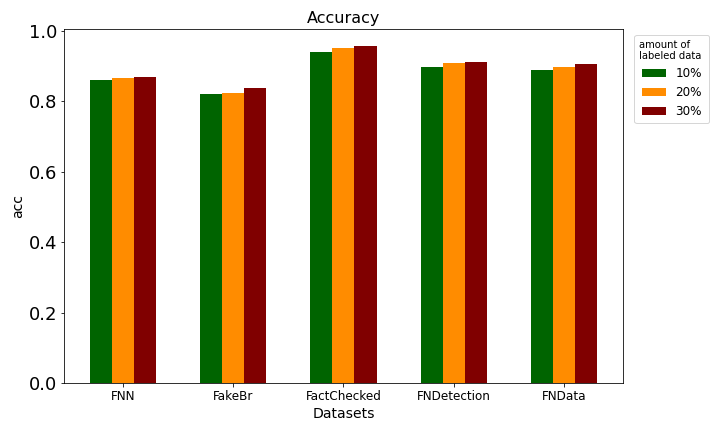}}
	\caption{Macro-F1 (a) and accuracy (b) comparision, for different amounts of labeled data}
	\label{fig:amount_of_labeled}
\end{figure}
As observed in \autoref{fig:amount_of_labeled},  the NOL-GAT model demonstrates improved performance across all datasets as the amount of labeled data increases. However, the extent of this improvement varies depending on the characteristics of each dataset. In smaller datasets, increasing the amount of labeled data has had a more significant impact on enhancing the model's accuracy and its ability to distinguish between fake and real news. In contrast, in larger datasets, the model has already achieved a reasonable performance even with a small amount of labeled data, and further increases in labeled data have had a more limited effect on the results.
The distribution of fake and real news also plays a crucial role in the model’s performance. In datasets with a relatively balanced distribution of both classes, the model initially exhibits stable performance, and improvements due to additional labeled data are more gradual. On the other hand, in datasets where the ratio of fake to real news is imbalanced, increasing the amount of labeled data has a more pronounced effect on improving the evaluation metrics. This suggests that the model requires more labeled data to enhance its performance in scenarios where the data distribution is skewed.
The trends observed for both accuracy and macro-F1 are generally similar. However, in some datasets, particularly those with imbalanced class distributions, the improvement in macro-F1 has been more pronounced than in accuracy. This indicates an increased ability of the model to differentiate between classes more effectively. These findings highlight that the NOL-GAT model performs well in semi-supervised settings, but the impact of labeled data on its performance depends on the dataset characteristics such as size and class distribution.
\subsubsection{Analysis of the impact of parameter k}
\label{sec:k}
\begin{figure}[!t]
	\centering
	\subfigure[]{\includegraphics[width=0.45\textwidth]{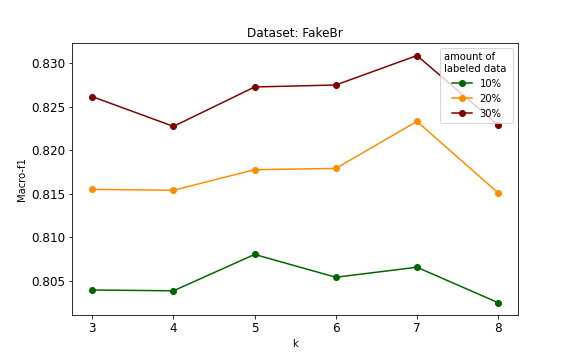}}
	\subfigure[]{\includegraphics[width=0.45\textwidth]{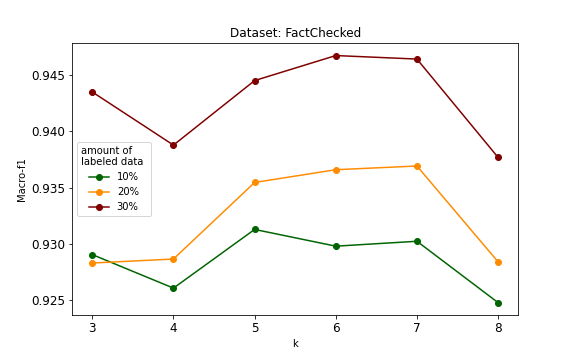}}
	\subfigure[]{\includegraphics[width=0.45\textwidth]{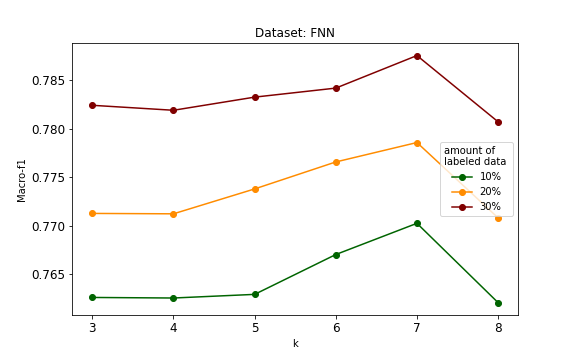}}
	\subfigure[]{\includegraphics[width=0.45\textwidth]{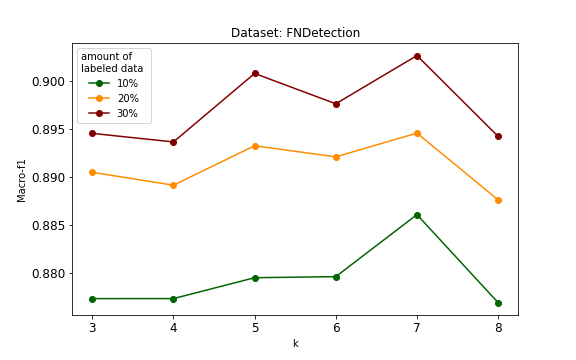}}
	\subfigure[]{\includegraphics[width=0.45\textwidth]{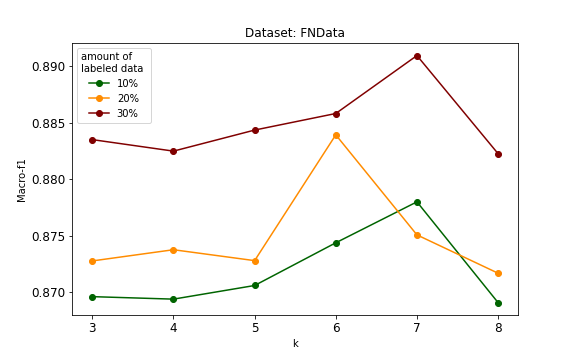}}
	\caption{Parameter sensitivity of k (number of nearest neighbors in $G_{KNN}$).
		(a): Fake.Br,
		(b): Fact-checked News,
		(c): FNN,
		(d): FakeNewsDetection,
		(e): FakeNewsData.}
	\label{fig:k}
\end{figure}

In this section, the effect of the k parameter in the KNN-based graph construction process on the performance of the NOL-GAT model is evaluated across five datasets.
The macro-F1 score, which reflects the model's ability to distinguish between different classes, is analyzed for various values of k.
The results indicate that the optimal value of k is generally within the range of 6 to 7.
At lower values of k (particularly k=3 and k=4), the macro-F1 score is lower, likely due to insufficient connectivity between nodes in the graph, resulting in a loss of important structural information.
This limitation reduces the model's ability to learn meaningful relationships within the data and ultimately impaired overall performance. 
As k increased to 6 and 7, the macro-F1 score reaches its peak, indicating that the graph connections are optimized in this range, allowing the model to achieve its highest level of class distinction.
However, in certain cases, a value of 5 also provides relatively good results.
On the other hand, setting k=8 leads to a performance drop across all datasets.
This decline may be attributed to an excessive number of graph edges, which blurred the distinction between different data points and introduced noise into the node relationships.
As a result, the model may have merged key features with irrelevant data, ultimately reducing its ability to accurately classify instances.
These observations underscore the importance of carefully selecting k during graph construction, as it can significantly enhance the model's performance.
\section{Conclusion}
\label{sec:conclusion}

With the increasing spread of misinformation and fake news in the digital age, developing efficient and accurate methods for detecting such content has become more critical than ever.
This paper introduced a novel model called NOL-GAT to address this challenge.
By employing an innovative approach for the dynamic and targeted selection of neighborhood orders in graphs, the model effectively overcame common limitations in graph neural network architectures, such as the inability to utilize information from distant neighbors and the over-squashing problem.
Experimental results demonstrated that NOL-GAT outperforms baseline models in key evaluation metrics such as accuracy and F1-score, especially in scenarios with limited labeled data. This highlights the model's ability to enhance information flow and reduce computational complexity, making it a practical choice for real-world applications.
By offering a flexible and scalable approach for analyzing graph-structured data, the NOL-GAT model can play a significant role in combating the spread of fake news.
This research not only introduces a new method for graph analysis but also serves as inspiration for further studies and developments in fake news detection and related applications. Future work could focus on employing advanced optimization techniques and integrating natural language processing methods to further enhance the potential of this model.


\bibliographystyle{elsarticle-num} 
\bibliography{references}

\end{document}